\documentclass[10pt]{article}
\usepackage{graphicx}
\usepackage{amsmath}
\usepackage{float}
\usepackage{amsfonts}
\usepackage[inner=0.75in, outer=0.75in, top=1in, bottom=1in, paperheight=11in, paperwidth=8.5in]{geometry}

\usepackage{parskip}
 \usepackage{nopageno}

 \usepackage{subcaption}
 
\usepackage{longtable}
\usepackage[table, x11names]{xcolor}
\definecolor{colour}{cmyk}{0.42,0.33,0,0}
\usepackage{multicol}
\usepackage{multirow}
\usepackage{comment}

\usepackage{tikz}
\usetikzlibrary{shapes, arrows, positioning, fit, backgrounds}
\usetikzlibrary{intersections}

\newcommand{\etal}{\textit{et al.} }

\newcommand{\ie}{\textit{i}.\textit{e}.}

\usepackage{bm}

\usepackage{authblk}

\begin{document}
	
	\title{\bf Better together: Using multi-task learning to improve feature selection within structural datasets}
	\author[1]{S.C.\ Bee}
	\author[1]{E.\ Papatheou}
	\author[1]{M.\ Haywood-Alexander}
	\author[1]{R.S.\ Mills}
	\author[2, 3]{L.A.\ Bull}
	\author[1]{K.\ Worden}
	\author[1]{N.\ Dervilis}
	\affil[1]{Dynamics Research Group, Department of Mechanical Engineering, University of Sheffield, Mappin Street, Sheffield S1 3JD, UK}
	\affil[2]{The Alan Turing Institute, The British Library, London, NW1 2DB, UK}
	\affil[3]{Department of Engineering, University of Cambridge, United Kingdom, UK, CB3 0FA}

	\date{}
	\maketitle
	\thispagestyle{empty}

	\section*{Abstract}
There have been recent efforts to move to population-based structural health monitoring (PBSHM) systems. One area of PBSHM which has been recognised for potential development is the use of multi-task learning (MTL); algorithms which differ from traditional independent learning algorithms. Presented here is the use of the MTL, ``Joint Feature Selection with LASSO'', to provide automatic feature selection for a structural dataset. The classification task is to differentiate between the port and starboard side of a tailplane, for samples from two aircraft of the same model. The independent learner produced perfect F1 scores but had poor engineering insight; whereas the MTL results were interpretable, highlighting structural differences as opposed to differences in experimental set-up. 

	\textbf{Key words: Automatic feature selection; generalisation; multi-task learning (MTL); sparsity; structural health monitoring (SHM).}
	
	
	\section{Introduction}
	
Population-based structural health monitoring (PBSHM) seeks to use a data-driven approach which considers multiple structures within one statistical model, as opposed to each structure individually. Multi-task learning (MTL,) is a suite of methods that considers multiple tasks simultaneously, as shown in Figure \ref{fig:PBML}. 

\tikzstyle{block} = [rectangle, draw, fill=blue!20, 
    text width=1.5em, text centered, rounded corners, minimum height=1.5em]
\tikzstyle{cloud} = [rectangle, draw, fill=red!20, 
    text width=1.5em, text centered, rounded corners, minimum height=1.5em]
\tikzstyle{process} = [rectangle, draw, fill=green!20, 
    text width=1.5em, text centered, rounded corners, minimum height=1.5em] 
\tikzstyle{process2} = [rectangle, draw, fill=green!20, 
    text width=2em, text centered, rounded corners, minimum height=1.5em] 
\tikzstyle{line} = [draw, -latex']

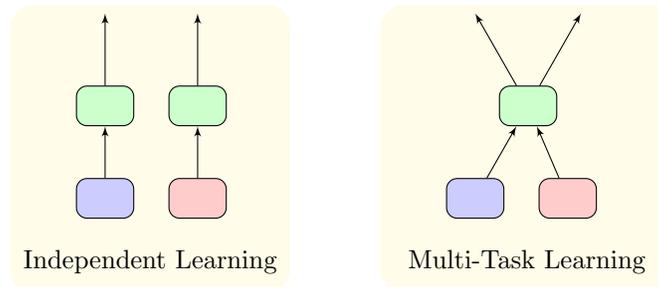
\begin{figure}[h]
\caption{Comparison of two types of machine learning.}
\centering
\label{fig:PBML}

\begin{tikzpicture}[node distance = 3.5em, auto]
    \node [block] (input1) {};
    \node [process, above of=input1] (process1) {};
    \node [cloud, right of=input1] (input2) {};
    \node [process, right of=process1] (process2) {};
    
    \coordinate[above of=process1] (output1);
    \coordinate[above of=process2] (output2);
    \coordinate[below left= of input1] (PH1);
    \coordinate[right of=output2] (PH2);
    
    \begin{scope}[on background layer]
    \node[fill, rounded corners=2ex, fill=yellow!10, fit=(PH1) (PH2), inner xsep=0mm, label={[label distance=-2em]270:Independent Learning}]{};
    \end{scope}
    
    \path [line] (input1) -> (process1);
    \path [line] (process1) -> (output1);
    \path [line] (input2) -> (process2);
    \path [line] (process2) -> (output2);
    
    \coordinate[right of=input2] (cont3);
    \coordinate[right of=cont3] (cont4);
    \node [block, right of= cont4] (input13) {};
    \node [cloud, right of=input13] (input23) {};
    \node [process, above of=input13, xshift= 2em] (process13) {};

    \coordinate[above of=process13, xshift= -2em] (output13);
    \coordinate[above of=process13, xshift= 2em] (output23);
    \coordinate[below left= of input13] (PH13);
    \coordinate[right of=output23] (PH23);
    
    \begin{scope}[on background layer]
    \node[fill, rounded corners=2ex, fill=yellow!10, fit=(PH13) (PH23), inner xsep=0mm, label={[label distance=-2em]270:Multi-Task Learning}]{};
    \end{scope}
    
    \path [line] (input13) -> (process13);
    \path [line] (process13) -> (output13);
    \path [line] (input23) -> (process13);
    \path [line] (process13) -> (output23);

\end{tikzpicture}
\end{figure}

Intuitively, training tasks together allows data to be shared, increasing the training dataset, and hence enabling the model to focus on meaningful \emph{feature selection}. Selection of appropriate features can reduce the signal processing requirements on the measured data \cite{Staszewski2002}. 

In previous SHM-related research \cite{Worden2003, Bull2021c}, initial feature selection was a manual selection of a frequency window or a greedy exercise, which was then fine tuned with further processing. However, ideally initial feature selection should be automatic to remove biases, but still be backed up with engineering insight. A study using a genetic algorithm (GA), as an automatic means of feature selection \cite{Worden2008}, improved classification when compared to an earlier study \cite{Worden2009}. However, GAs are greedy algorithms and also do not inherently provide \emph{sparse} solutions. To reduce signal processing requirements and aid engineering insight, a sparse solution which utilises only information-rich features is desirable and can aid the selection of more interpretable features in engineering practice. 

Arguably the most popular method to generate features is Principal Component Analysis (PCA) \cite{Wang2009, Dackermann2014, Gordan2017}. PCA reduces the dimensionality of the dataset whilst maximising the preservation of information \cite{Jollife2016}. However, PCA does not enforce a reduction in the number of original features, it only maximises variance between the ``new'' features. Whilst PCA is common for statistical applications, there may be the opportunity to select features based on physics as well.

The statistical model presented here utilises MTL for both \emph{automatic feature selection} and classification. In addition to automatic feature selection, the model presented also encourages sparsity. The majority of feature selection methods used in SHM are actually forms of \emph{feature extraction}, which do not encourage sparse selection, so not only does this work demonstrate a sparse solution (for independent learning and for multi-task learning), but it can also be shown to improve the selection of meaningful features in SHM tasks. 

Section \ref{section:Background} will discuss the background of the algorithm, Section \ref{section:dataset1} details the dataset and the results from the application of the algorithm and, finally, Section \ref{section:conclusion} concludes this paper.

	\section{Background}
	\label{section:Background}
The following section will detail the set-up of the model used in this research, how it is solved and how the success of the algorithm is measured. 

	\subsection{The Loss Function}
	\subsubsection{Empirical Loss}
For both the independent learner and MTL, \emph{logistic regression} is used for classification. The output of the classification is either True or False. To mathematically model ``True" or ``False", a linear regression is used in the form of ``$\bm{W}^T\bm{x}^{(i)}$'', followed by an activation function, such as the sigmoid function, to generate a predicted value between (0,1),
\begin{equation}
 \hat{y}^{(i)} = \frac{1}{1 + e^{-\bm{W}^T\bm{x}^{(i)}}}
 \end{equation}

where $\bm{x}^{(i)} \in \mathbb{R}^M $ is an observed set of readings of $M$ features, $\bm{W} \in \mathbb{R}^M$ is the weight vector with a corresponding weight for each of the $M$ features, and the superscript $i$ refers to one of the $N$ observed sets of readings (\ie, $i \in \{1, 2, ..., N\}$). 

It is likely that when the sigmoid function is used, the value of $\hat{y}^{(i)}$ will not be 0 or 1, rather a value in between. If this is the case, then the value can be thought of as the probability that the result is True. To determine the classification output, a threshold value is chosen as $0.5$.

The loss function is dependent on the predicted value $\hat{y}$ and the observed value $y$, 
\begin{equation}
\label{eq:lossFun}
 J(\hat{y}^{(i)}, y^{(i)}; \bm{W}) = -\frac{1}{N} \sum^N_{i=1} \left( y^{(i)}log(\hat{y}^{(i)} )+(1-y^{(i)})log(1-\hat{y}^{(i)} ) \right)
 \end{equation}
 
This form of the loss function is known as the \emph{empirical loss function} and is a measurement of error per sample which is averaged over all $N$ measurement sets.

	\subsubsection{Independent Learner}
With only the empirical loss, a model will be prone to over-fitting and hence a further term is added to the loss function to prevent over-fitting. The Least Absolute Shrinkage and Selection Operator (LASSO) algorithm \cite{Tibshirani1996} adds a regularisation term to the empirical loss function (equation (\ref{eq:lossFun})) in the form of an $\ell^1$ norm. To understand the impact of the $\ell^1$ norm on the loss function, it is useful to understand the general form of $\ell^P$ norms. If $\bm{W}$ is a vector, $[w_1, w_2, ... w_M ]$, then the $\ell^P$ norm is given by:
\begin{equation}
||\bm{W}||_P = \left( \sum^M_{j=1} w_j^P \right)^\frac{1}{P}
\end{equation}

The optimal solution for a set of training data is given by minimising the empirical loss, equation (\ref{eq:lossFun}). This solution would be over-fitted and generalise poorly, such that although the solution fits the training data well, it would not fit a new dataset well. In order to provide generalisation, the LASSO adds an ($\ell^1$) penalty, such that the resulting constraint is $\sum^M_{j=1}|w_j| < c$.  By implementing regularisation the resulting model is more likely to be representative of the new dataset. However, it should be noted that if the regularisation dominates, then under-fitting will occur, and the model will perform poorly for both the training and the test datasets. The total loss function given by the LASSO algorithm is,
\begin{equation} \label{lassoLossFunc}
\Gamma(\hat{y}^{(i)}, y^{(i)}; \bm{W}) = J(y^{(i)}, \hat{y}^{(i)}; \bm{W})+ \lambda \|\bm{W}\|_1
\end{equation}
where $\lambda$ is a scalar known as the regularisation parameter, and $\|\bm{W}\|_1 = \sum^M_{j=1}|w_j|$ is the $\ell^1$-norm of the weight vector.

This total loss function has the ability to increase sparsity as it applies a constraint using the $\ell^1$ norm.
	
	\subsubsection{Multi-task Learner}
To use the loss function across multiple tasks within one model, the loss function must consider all of the tasks. Joint regularisation with LASSO was introduced in MTL by Obozinski \etal\cite{Obozinski2006} to encourage features to share the same sparsity pattern among similar tasks by adding an $\ell^{2, 1}$ constraint. The constraint is $\ell^2$ across the different tasks, which is then combined into $\ell^1$ across the features, 
\begin {equation} \label{MTLConstraints}
\begin{aligned}
\|\bm{W}_{L}\|_2 &= \left(\sum^L_{l=1} w_{j,l}^2\right)^{1/2}  &&\text{($\ell^2$ across the tasks)}\\
\|\bm{W_M}\|_{2, 1} &= \sum^{M}_{j=1} |\left(\sum^L_{l=1} w_{j,l}^2\right)^{1/2}| &&\text{($\ell^1$ across the features)}
\end{aligned}
\end{equation}
where $L$ is the number of tasks, and $\bm{W}_{L} \in \mathbb{R}^M$.

For each feature, there is an $\ell^2$-norm constraint between the tasks. The next layer of constraint is the $\ell^1$-norm. The application of the $\ell^1$-norm means that sparsity is encouraged across the features. If a feature has a zero weight with this constraint, then all of the tasks will have a zero weight for the given feature. When all of the tasks have zero weights for the same features, then the tasks will share the same sparsity pattern.

With multiple tasks the empirical loss function (equation (\ref{eq:lossFun})), becomes,

\begin{equation}
\label{eq:lossFunMTL}
 J(\hat{y}^{(i)}_l, y^{(i)}_l; \bm{W}_l) =  -\frac{1}{L}\sum^{L}_{l=1}\frac{1}{N_l} \sum^{N_l}_{i=1} \left( y^{(i)}_llog(\hat{y}^{(i)}_l )+(1-y^{(i)}_l)log(1-\hat{y}^{(i)}_l)  \right)
 \end{equation}
where $\bm{W}_l = [w_{1, l}, w_{2,l}, ... w_{M,l}]$ refers to the weight vector, and $N_l$ refers to the number of samples for a given task $l$. 

Using the empirical loss function defined above (equation (\ref{eq:lossFunMTL})), the total loss function (equation (\ref{lassoLossFunc})), becomes,

\begin{equation} \label{MTLLassoLossFunc}
\Gamma(\hat{y}^{(i)}_l, y^{(i)}_l; \bm{W}_l) = J(\hat{y}^{(i)}, y^{(i)}; \bm{W_l}) + \lambda \|\bm{W}_M\|_{2, 1}
\end{equation}
where $|\bm{W}_M\|_{2, 1} = \sum^{M}_{j=1} |\left(\sum^L_{l=1} w_{j,l}^2\right)^{1/2}|$.

	\subsection{Solving with Gradient Boosting}
To enable further sparsity, both in independent learning and in a multi-task learning setting, ``gradient boosting'' can be used to solve the algorithm. \emph{Gradient boosting} is a form of co-ordinate descent such that, rather than updating all of the weights in each iteration, one weight is updated and the rest of the weights remain constant. By only updating one weight at a time, weights that have low influence on the loss function will remain unchanged from their original zero value; hence, a solution with increased sparsity can be achieved.  

Forward steps are taken to reduce the total loss (equations (\ref{lassoLossFunc}) and (\ref{MTLLassoLossFunc})), and backward steps are taken to reduce the empirical loss (equations (\ref{eq:lossFun}) and (\ref{eq:lossFunMTL})) \cite{Zhao2004}. Gradient boosting has a hyperparameters  $\epsilon$, which is the increment that the weight is changed by, and $\xi$, which is the minimum loss required to complete an iteration. When these parameters take larger values, the resulting matrix is more likely to be sparse, as only the features with significant impact on the loss will result in updated weights; the larger these parameters get, the less accurate the algorithm is, as the values of the weights are less sensitive. As the values of these hyperparameters reduce, the simulations are likely to take longer to run, and potentially result in less-sparse solutions; however, the results are likely to have increased accuracy.
	
	\subsection{Algorithm Performance}
To evaluate the success of the algorithm during training, so that the optimal hyperparameters are selected, it is necessary to define how the performance of the algorithm is evaluated. There are two elements to this: how accurate the algorithm is at correctly predicting the class and how sparse is the weight matrix to generate the predicted class.

Often to determine the performance of an algorithm during the training stage, the value of the loss function is used. However, when comparing the losses of models with different hyperparameters, the value is not intuitive and is not comparable across multiple tasks and multiple datasets. Instead, a performance indicator which is bounded to a range (e.g. 0-100\%) may hold more meaning. For the accuracy, the F1 score is used here.

A large number of features is often not interpretable, therefore, sparsity is a desirable characteristic as, not only does it offer a reduction in processing requirements, but it also has the additional benefit of being interpretable.  To measure sparsity the Gini Index is used. Hurley and Rickard \cite{Hurley2009}, provide a comparison of different measures of sparsity, they recommend that the Gini Index is the most robust measure. The Gini Index was originally used to measure the distribution of wealth \cite{Farris2010}; however, it has been used across multiple disciplines including ecology \cite{Cordonnier2015}, medicine \cite{Bandara2022} and in 2017, the Gini Index was introduced in encoder-based applications to assess the health condition of rotating machinery \cite{Zhao2018}. Hurley and Rickard's comparison \cite{Hurley2009}, defines the Gini Index as,
\begin{equation} \label{Gini}
G (\bm{W})= 1- 2\sum^M_{j=1} \frac{w_j}{||\bm{W}||_1}\left( \frac{M-j+\frac{1}{2}}{M}\right)
\end{equation}

where the $\bm{W}$ vector used in equation (\ref{Gini}) has been ordered by magnitude such that $|w_x| < |w_{x+1}|$ for $x \in \{1, ..., M\}$.

The Gini Index is beneficial, as not only does it consider whether the weights have been activated or not, but also the distribution of magnitude of the weights; this measure gives a result $\in [0, 1]$.
	
	\section{Application 1: Piper Aircraft Tail Plane}
	\label{section:dataset1}
 To determine whether Joint Feature Selection using the LASSO algorithm will improve the classification accuracy, a subset of the tail plane dataset used by Bull \etal \cite{Bull2021c} will be analysed. In the paper, the data was used to determine whether domain adaptation could be used to transfer learning between structures and hence improve novelty detection. One of the areas for future work highlighted in the paper was automatic feature selection. The dataset will be used in a way to test the \emph{concept} of the MTL LASSO algorithm with a structural dataset. The classification task is synonymous with analysing structures pre-repair and post-repair. The concepts and discussion herein can be further researched and applied to SHM problems. 
 
	\subsection{Initial Dataset and Dataset Generation}
The dataset used is made up of two tail-planes from a PA-28 `Arrow’ aircraft which are labelled A and B. The elevators and wing tips were removed from the tail-planes and then each tail-plane was cut in half to create a port and a starboard side. Tail-plane A and B have more or less the same geometry (although B was cut asymmetrically). The tail-planes have come from the same aircraft model, which would make them a homogeneous population. A population is classified as ``homogeneous'' if the structures are ``nominally identical'' \cite{Gosliga2022} and there is ``structural equivalence'' \cite{Gosliga2021b}. As the tail-planes have been cut asymmetrically, the tail-planes from A and B are no longer strictly homogeneous (identical); rather they are classed here as ``weakly homogeneous''. The classification task is to determine from the frequency response function (FRF), which of the datasets is from the A tail-plane and which is from the B tail-plane. This classification would be synoynmous with reviewing a structure in normal condition (A tail plane) and damaged condition (B tail plane), or vice versa. The purpose of this classification task is to investigate the suitability of the algorithm. As there are port and starboard parts of the tail-plane, there are two classification tasks. 

\begin{figure}[!h]
\centering
\begin{subfigure}{.5\textwidth}
  \centering
  \includegraphics[width=0.95\linewidth]{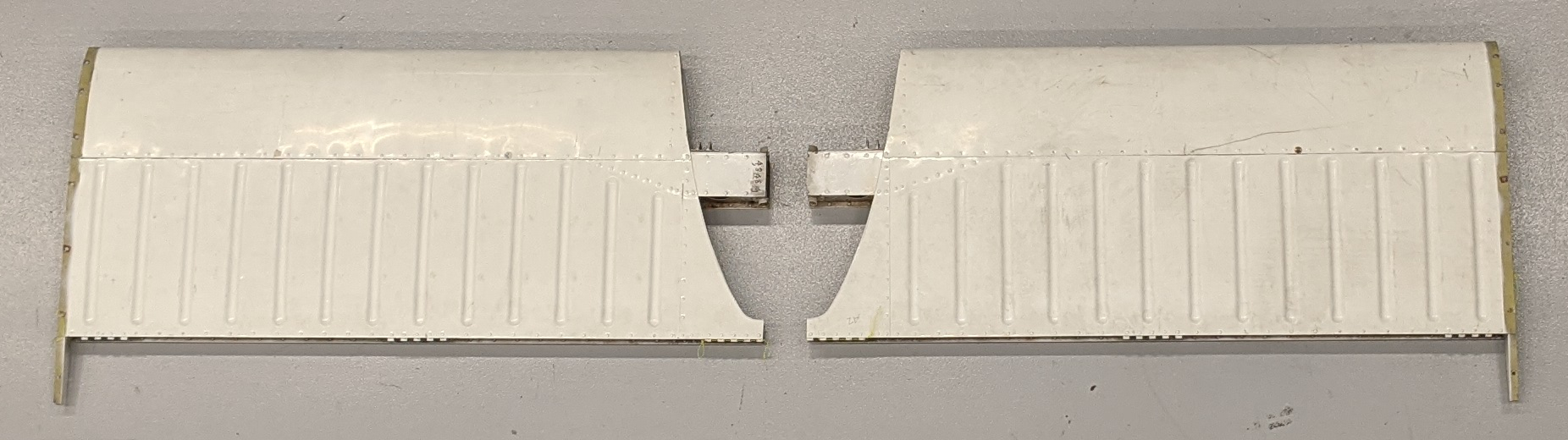}
  \caption{A1 (port) and A2 (starboard)}
\end{subfigure}%
\begin{subfigure}{.5\textwidth}
  \centering
  \includegraphics[width=0.95\linewidth]{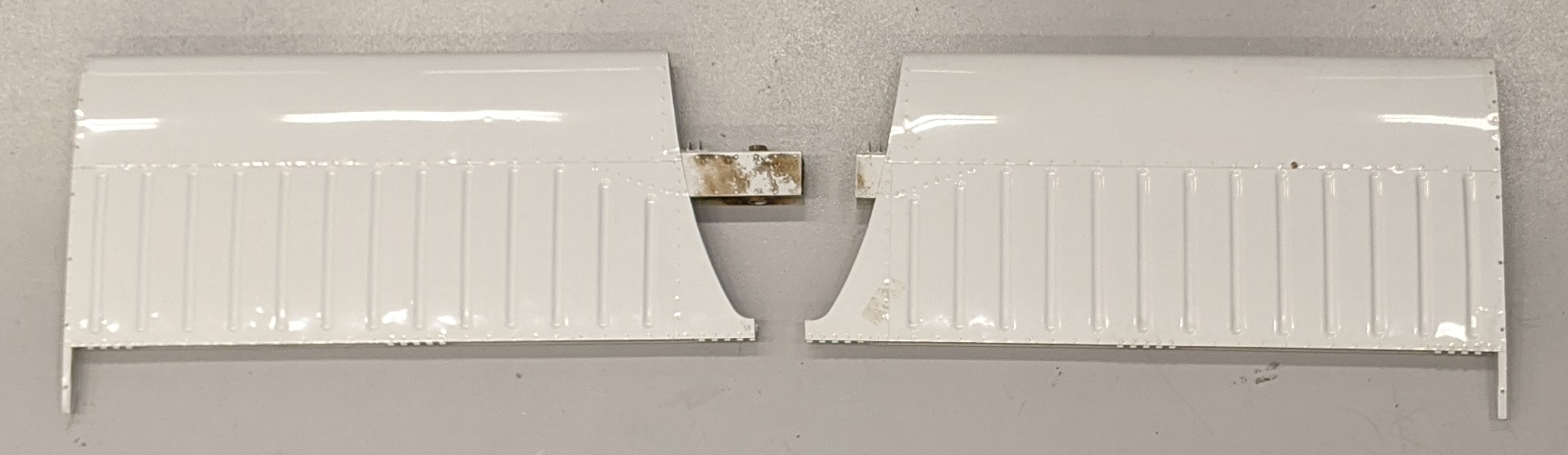}
  \caption{B1 (port) and B2 (starboard)}
\end{subfigure}
\caption{The tail-plane structures.}
\label{fig:TailPlanePhotos}
\end{figure}

Gaussian white noise excitation was applied to each of the tail-planes at a frequency bandwidth of 1kHz and a resolution of 0.3125Hz, this resulted in 3200 points in the frequency spectrum. The analysed frequency range was between 33.75 and 217.1875Hz (points 107 to 695), as this is deemed the useful range of the FRF. 

\begin{figure}[ht]
\centering
\includegraphics[width=0.9\linewidth]{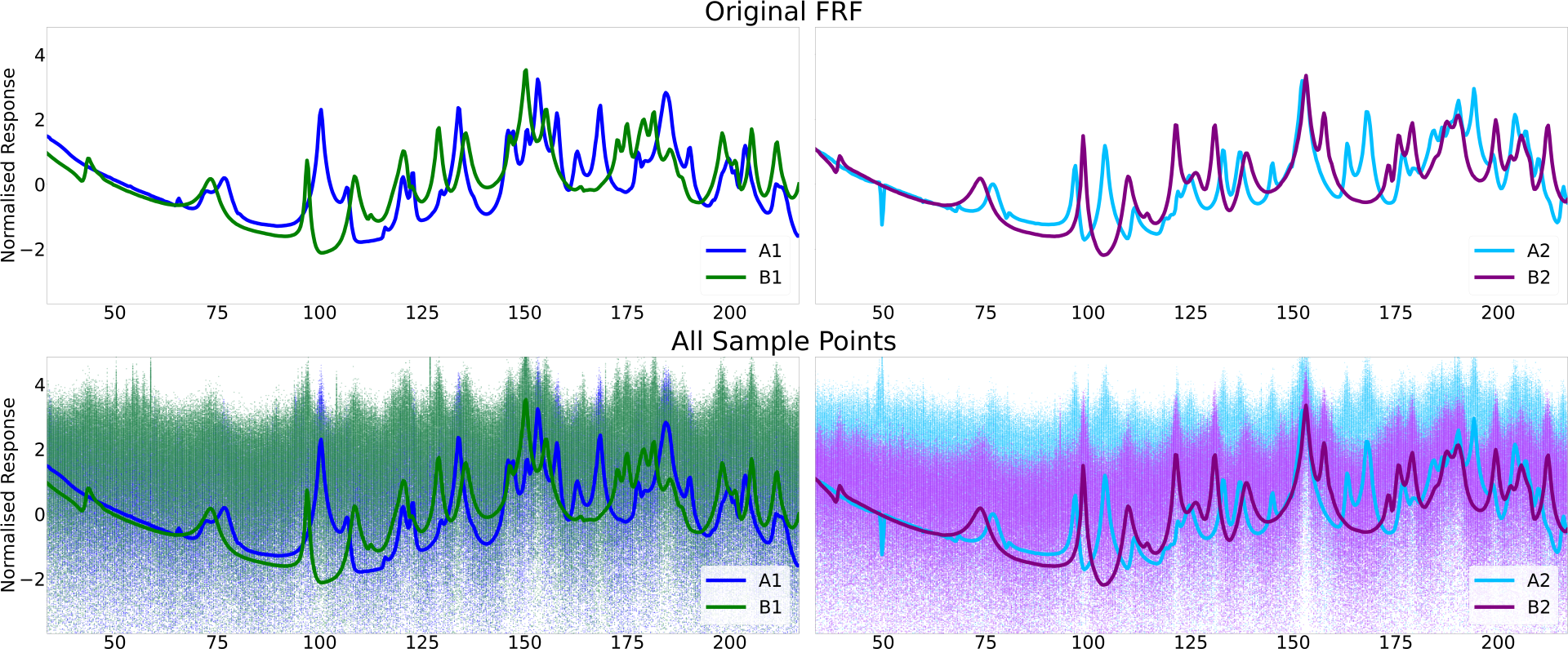}
\caption{\emph{Top:} The frequency response of tail-planes A and B. Showing port measurements (Task A, \emph{left}, blue for A and green for B) and starboard measurements (Task B, \emph{right}, sky blue for A and purple for B), based on the average response of the 180 measurement points. \\  \emph{Bottom:} The frequency response of tail-planes A and B (as in \emph{top}) and all 750 sample points per structure shown as very small translucent points (Task A, \emph{left} and Task B, \emph{right}; with Class 1, blue for Task A and sky blue for Task B, and Class 2, green for Task A and purple for Task B).}
\label{fig:full_FRF}
\end{figure}

The response is not measured in one location on the structure, rather, each of the port and starboard sections of tail-plane A and B had 180 measurement response points. The normalised FRF can be generated by averaging the measured response across the 180 points, and then normalising (same as the methodology used in \cite{Bull2021c}). The resulting FRF can be seen in the top row of Figure \ref{fig:full_FRF}. 

There are 588 measured frequencies between 33.75 and 217.1875Hz, each of the frequencies corresponds to one feature. For each frequency, or feature, there is only one data point per feature as the data point is the summed average across all of the 180 sample points. To train a model requires $N$ sample points. Jain and Waller \cite{Jain1978} proposed that for uncorrelated features, the optimal sample size is $N=M+1$ (where $M$ is the total number of features), whereas for highly-correlated features the optimal sample size is proportional to $N=M^2$. This relationship is further backed up by research of Hua \etal\cite{Hua2005}. To increase the dataset from $1$ to $N$, a dataset was generated using Monte-Carlo sampling; with the measured data as a mean and the variance estimated using the coherence function \cite{Worden1998},
\begin{equation}
\sigma(H_p(\omega))= \frac{\sqrt{1-\gamma_p^2(\omega)}}{|\gamma_p(\omega)|\sqrt{2N}}H_p(\omega)
\end{equation}
where $H_p(\omega)$ is the measured frequency response at a given frequency, $\gamma_p(\omega)$ is the standard coherence function and $N = 6$ is the number of values used to compute the FRF (\ie{} the number of average values used to generate the resulting measured value). 

The data are assumed to have a Gaussian distribution and therefore random samples can be generated for each measured response point (10,000 here). The mean and standard deviation for each response point can be calculated and used to generate a further sample set which will be used in the analysis. The aim is to generate a sparse solution; however, a sparse solution would only be achievable if correlation between features existed. At this stage of the statistical model development, the effective number of meaningful features is unknown. 

It is anticipated that there will be correlation between the features and there is the potential for high correlation. High correlation between tasks would suggest that the number of samples required would be 354,744 ($588^2$, or 172,872 per class). A practical sample size of 1,500 per task (\ie{} 750 in each class) was generated for training and validation; much less than 354,744. To reduce the number of features analysed with the number of samples available, the 588 frequency measurements were split into 6 windows; each with 98 consecutive frequencies. This would reduce the suggested number of samples to 9,604 ($98^2$); the chosen sample size is still $\sim$6 times smaller than the recommended; however, sample size is now closer to the ``rule of thumb''. The number of windows is a hyperparameter that will require tuning, however, six windows will be used while training $\epsilon$ and $\xi$. 

To validate the trained model, five-fold cross validation was implemented. Five-fold cross validation is generally accepted as the lower end of the permissible values for k-fold cross validation \cite{Priddy2005, Mirkin2011}. It was implemented to reduce bias within the model and enable optimal values for the hyperparameters to be selected. A further 500 samples per task (250 per class) were generated for testing purposes. 

	\subsection{Hyperparameter training}
There are two sets of hyperparameters which need to be learned; firstly, those in the algorithm: the step size $\epsilon$, and tolerance $\xi$. Because of the curse of dimensionality, another hyperparameter is the number of windows that the FRF will be analysed over. 

Initially, the FRF was split into six windows such that there were 98 frequencies in each window. Three models were produced; an independent learner for Task 1, an independent learner for Task 2 and a mult-task learner for tsak 1 and 2. All three models were trained with the same hyperparameters, the values of the step size, $\epsilon$, were selected to be approximately three times smaller than the previous value,  $\epsilon\in\{ 1, 0.3, 0.1, 0.03\}$, and the tolerance was trialled for 3 different orders of magnitude, $\xi\in\{ 0.1, 0.01, 0.001\}$. All combinations of $\epsilon$ and $\xi$ are tried subject to $\epsilon > \xi$. 

\begin{figure}[p]
\centering
\begin{subfigure}{0.95\textwidth}
  \centering
  \includegraphics[width=\linewidth]{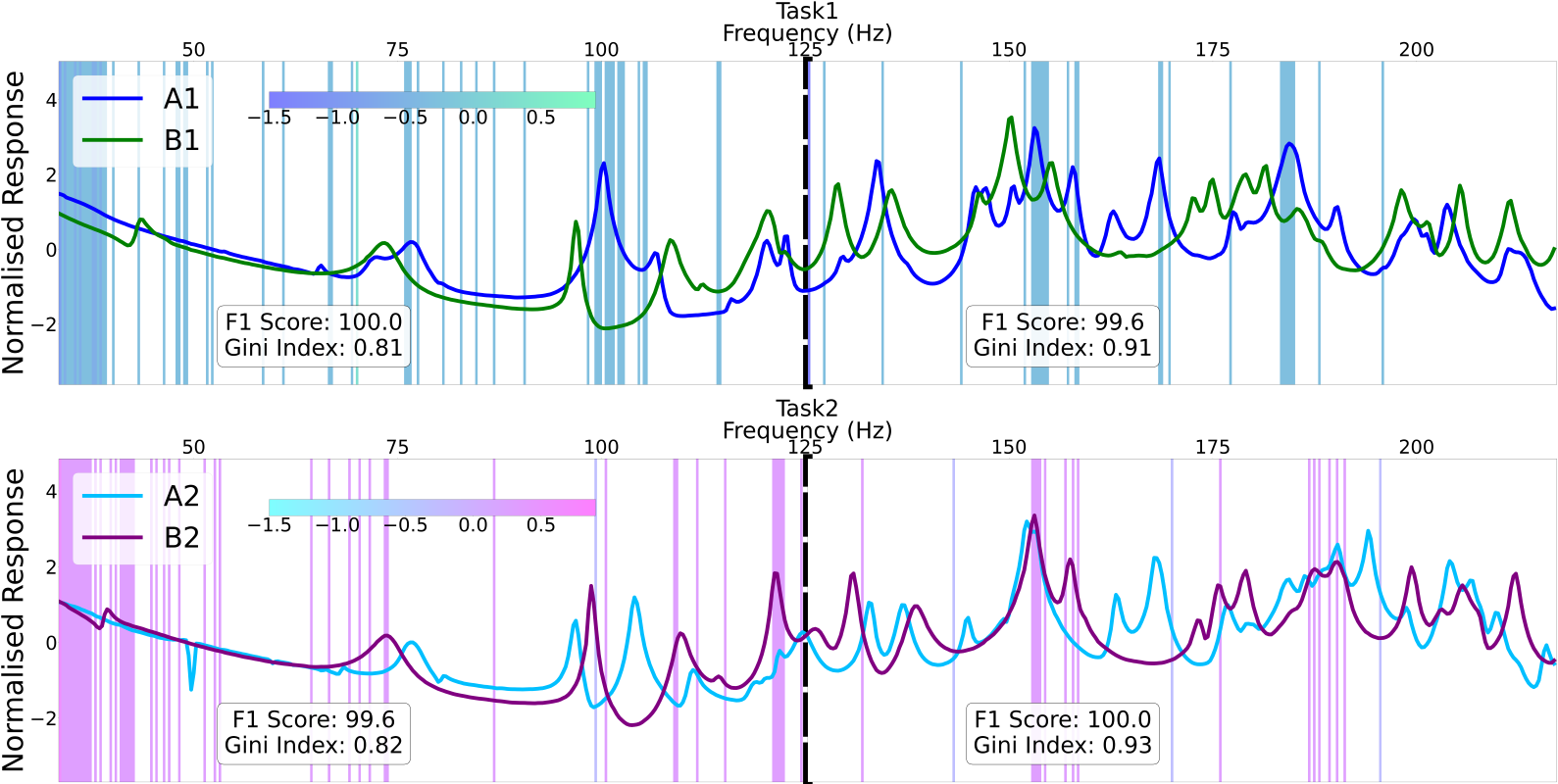}
  \caption{Independent Learner.}
\end{subfigure}
\begin{subfigure}{0.95\textwidth}
  \centering
  \includegraphics[width=\linewidth]{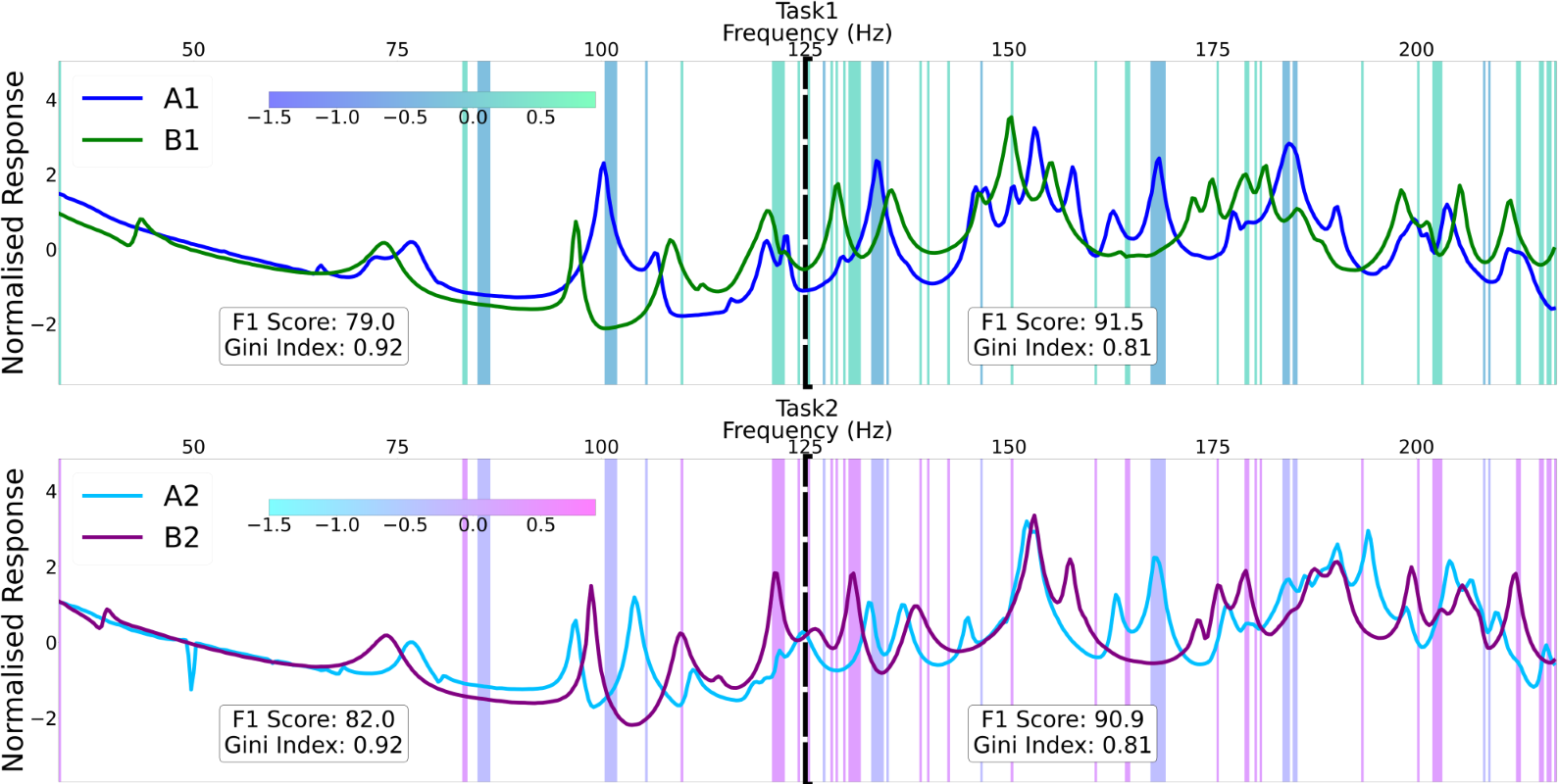}
  \caption{MTL.}
\end{subfigure}
\caption{FRF and activated weights for test data for both Task 1 (upper, Class 1 blue and Class 2 green)  and Task 2 (lower, Class 1 sky blue and Class 2 purple) for (a) LASSO and (b) Joint Feature Selection with LASSO, $\epsilon=0.3$ and $\xi = 0.01$. The activated weights are shown as vertical lines and the colour of the vertical line represents the value of the weight. The dashed black vertical line represents the boundary for the two windows.}
\label{fig:final}
\end{figure}

For 1500 samples in the training data set (75\% of the total 1000 samples per class), to achieve the optimal number of features for highly-correlated data \cite{Jain1978, Hua2005}, then 38 features per window would be required ($38^2 = 1,444$), or 16 windows in total. Therefore, the number of windows that should be analysed is from 1 to 16 windows. For simplicity, $\epsilon$ and $\xi$ are fixed at the values which were optimal for six windows. The best performing algorithms had 2 windows, following the optimisation for the number of windows, the value of $\epsilon$ was again optimised for $\epsilon\in\{ 0.3, 0.2, 0.1\}$.

	\subsection{Trained model comparison}
Following tuning of the hyperparameters, the resulting hyperparameters are: $N_{windows}=2$, $\epsilon=0.2$ and $\xi=0.01$ for both the independent and the multi-task models. Figure \ref{fig:final} shows the FRF overlaid onto vertical lines which represent the activated weights, and therefore the activated frequencies of the solution. 

The independent learners (Figure \ref{fig:final} (a)), both have a lot of activated frequencies at the lower end of the frequencies analysed. The large number of activated weights at this lower end is not informative of the differences in the structure, rather the differences in the rigid body movements created because of the experimental set-up. For Window 2 (frequencies $>$125Hz), there appear to be similarities in the activated weights between Task 1 and Task 2; this is to be anticipated, as the samples are from the same aircraft; this is also the motivation for using a multi-task learner on this dataset!

Figure \ref{fig:final} (a) shows that near-perfect F1 scores are obtained for both tasks and both windows. However, the Gini Index for Window One across both tasks is 10\% points lower than the Gini Index for Window Two. For Task 1, it could be argued that the feature set from either of the windows could be used, Window 1 has a perfect F1 score, however, there is a marked improvement in Gini Index in Window 2 which is arguably worth the small reduction in F1 score. The selection of the window for Task 2 would be Window 2 as it has a perfect F1 score and a high Gini Index.  

For Figure \ref{fig:final} (b), the F1 scores are lower than the independent learner equivalent. For Window One the frequencies that have been activated are at the higher end of frequency range, and there is only one frequency activated less than 80Hz. The absence of low frequencies being activated indicates that the experimental set-up (which would be different across all four samples), is no longer useful to differentiate between the two classes. Window Two outperforms Window One for both tasks, and the F1 scores are lower.

	\subsection{Transfer of Results}
 To determine the success of the models, it is useful to implement the weight matrix (features selected) from the two independent learners and the multi-task learner on the two existing tasks and on a third, unseen task. 

 \begin{figure}[ht]
\centering
\includegraphics[width=\linewidth]{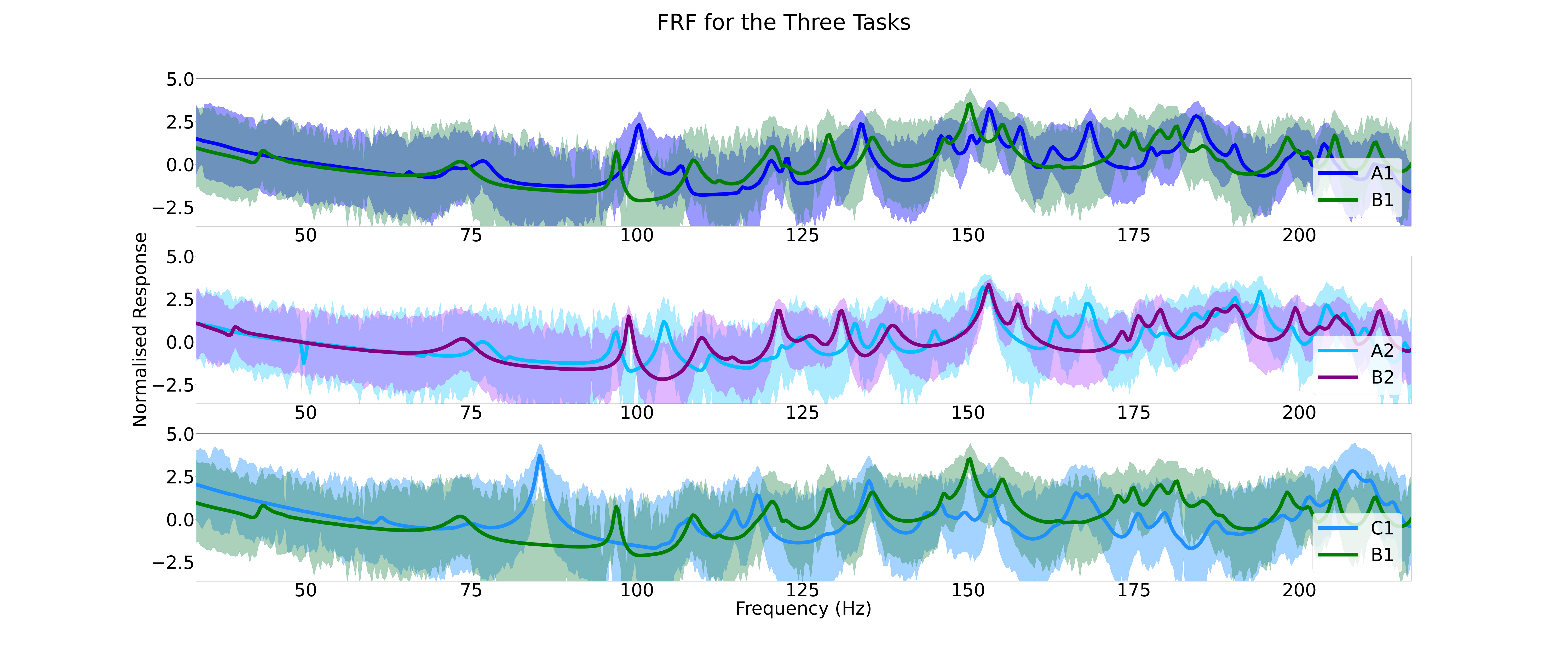}
\caption{FRF for the three different tasks showing plus and minus one standard deviation (shaded band). \emph{Top:} A1 vs B1, \emph{Middle:} A2 vs B2, and \emph{Bottom:} C1 vs B1}
\label{fig:3Tasks}
\end{figure}

 In the tail plane dataset used by Bull \etal \cite{Bull2021c} there is a third tail plane, C, which is taken from a PA-28 `Cherokee' aircraft. The structure is no longer homogeneous, as it is a different variant of aircraft. The starboard side of the tail plane (C2) was damaged and the port side of the tail plane (C1) was not damaged. To formulate a task which has similarities with the original task set, a useful classification is ``C1 vs B1''. Figure \ref{fig:3Tasks} shows the FRF for the two original tasks and the 

This experimentation has shown that the multi-task learner has generalised well over the two tasks and, from the generalisation, is focusing on frequencies which are likely to result in good information about the structures. 	

        \section{Application 2: GNAT Aircraft Wing}	
        \label{section:dataset2}

        \subsection{The Dataset}
To further determine whether Joint Feature Selection using the LASSO algorithm will improve engineering insights, a subset of the GNAT dataset used by Manson \etal \cite{Manson2003} will be tested.

\begin{figure}[ht]
\centering
\includegraphics[width=0.9\linewidth]{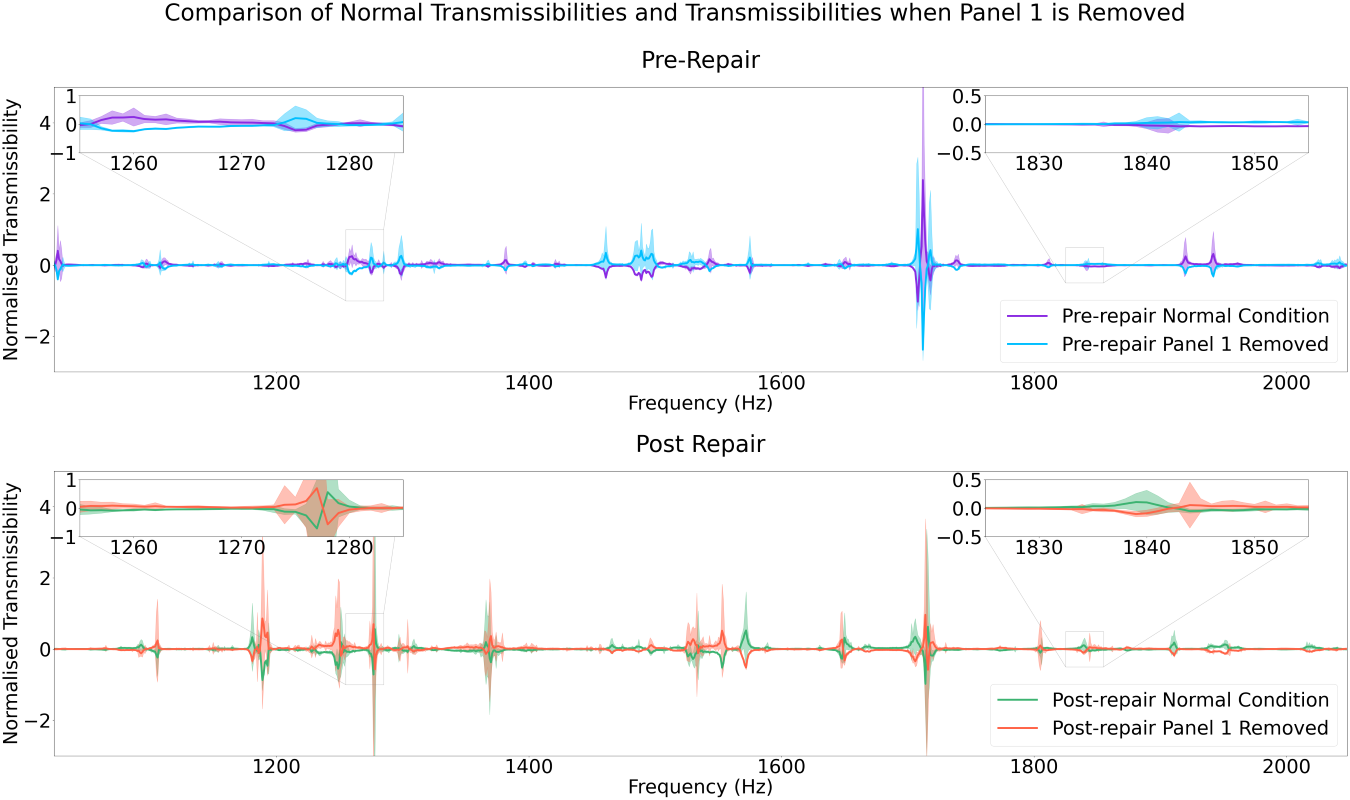}
\caption{\emph{Top:} The transmissibility pre-repair (Task A, \emph{top}) and post-repair (Task B, \emph{bottom}) of reference transducer AR to response transducer A1 for normal condition and panel 1 removed. Pre-repair, Task A, has class 1 as normal condition (purple) and class 2 as panel 1 removed (sky blue) and post-repair, Task B, also has class 1 as normal condition (green) and class 2 as panel 1 removed (orange). One standard deviation of banding is shown for each class. Inset graphs at 1255 to 1285Hz and 1825 to 1855 are shown on each graph to visualise the graph with increased magnification}
\label{fig:transmissibilities}
\end{figure}

        \subsection{Trained model comparison}

\begin{figure}[p]
\centering
\begin{subfigure}{0.95\textwidth}
  \centering
  \includegraphics[width=\linewidth]{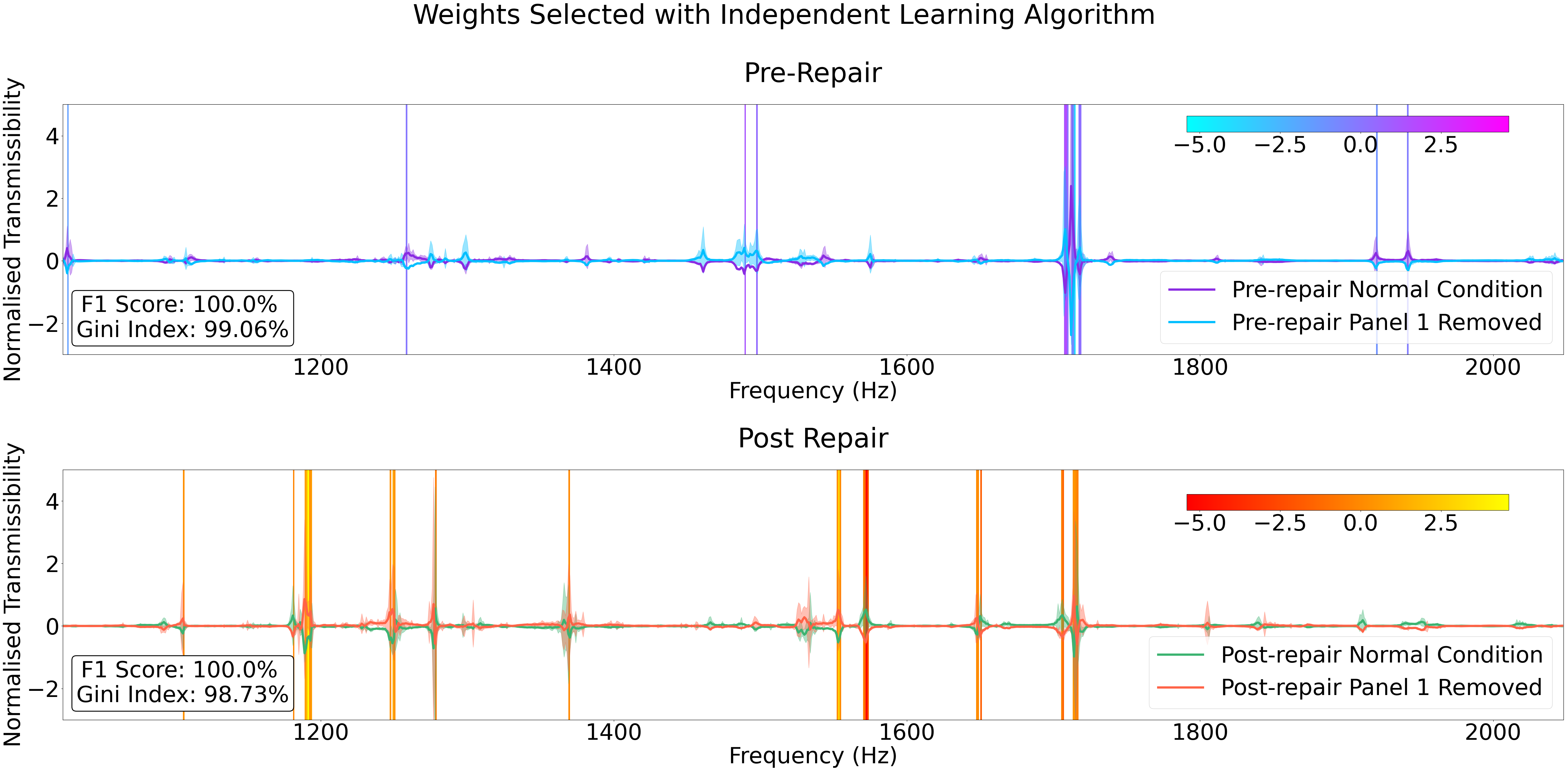}
  \caption{Independent Learner.}
\end{subfigure}
\begin{subfigure}{0.95\textwidth}
  \centering
  \includegraphics[width=\linewidth]{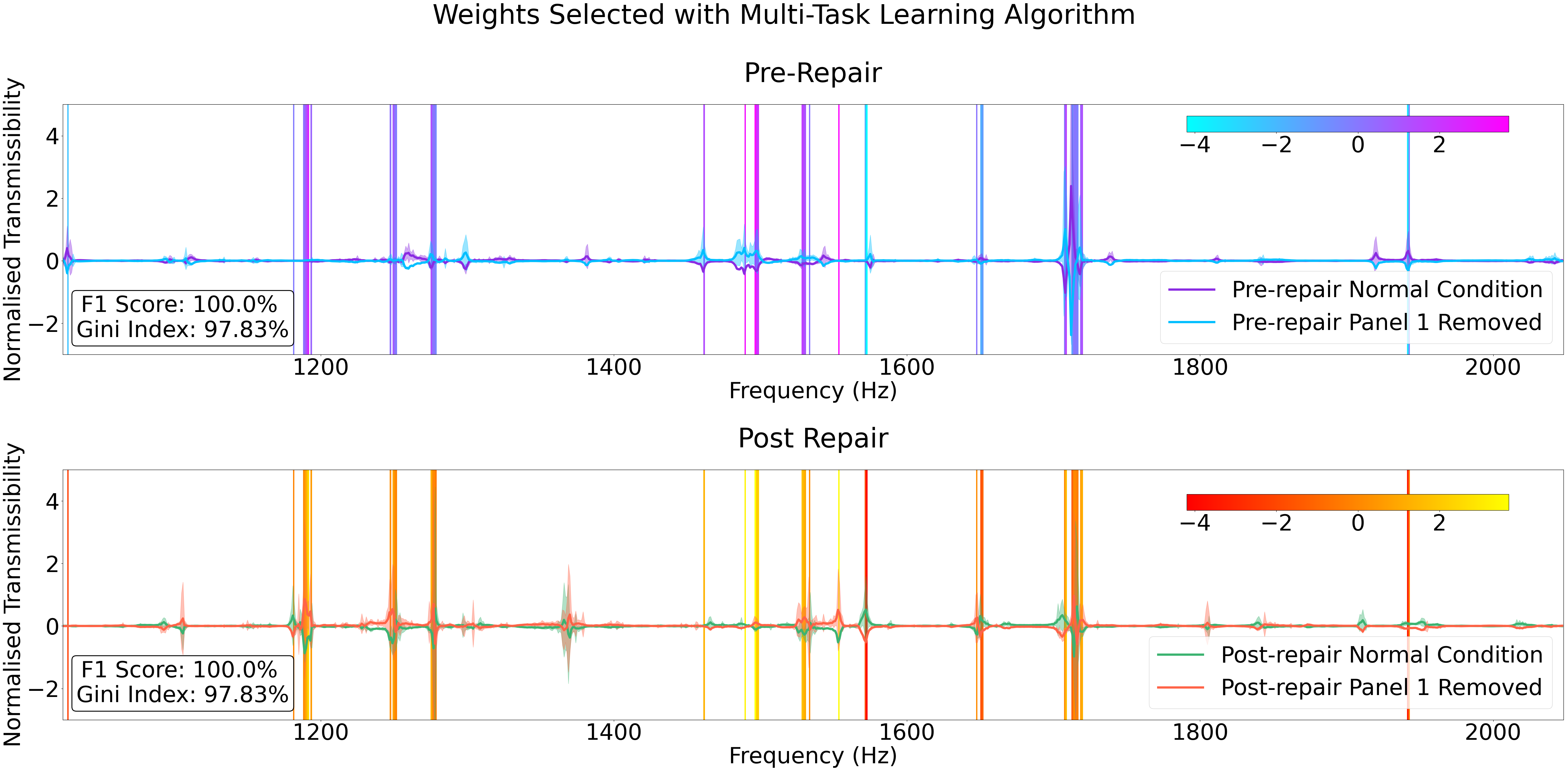}
  \caption{MTL.}
\end{subfigure}
\caption{Transmissibilities and activated weights for test data for both Task 1 (upper, normal condition purple and panel 1 removed blue)  and Task 2 (lower, normal condition green and panel 1 removed orange) for (a) LASSO and (b) Joint Feature Selection with LASSO, $\epsilon=0.1$ and $\xi = 0.1$. The activated weights are shown as vertical lines and the colour of the vertical line represents the value of the weight.}
\label{fig:final2}
\end{figure}

        \section{Concluding Remarks}
	\label{section:conclusion}
Successful automatic feature selection has been demonstrated on the tail-plane dataset. Gradient boosting yielded results with good sparsity in both the independent and MTL setting. 

Classification of the two tasks was perfect (\ie{} 100\% F1 score) for the independent learner; however, Window 1 shows that differences in experimental set-up was a key driver in feature selection. The features that were consequently selected were specific to the one task and class analysed in each model and did not generalise for the other task.

Compared to independent learning, and without any engineering judgement, the result of the MTL is such that the F1 score is lower, as is the Gini Index, and therefore the MTL has not outperformed the independent learner. However, the inference that can be taken from the final result is quite powerful. Although the metrics have underperformed in MTL, The activated weights in the MTL are more meaningful and representative of changes in the structure. 

In this work, MTL using LASSO has highlighted appropriate features, and this can be justified with visual inspection (engineering judgement). Further work is needed to make this technology applicable within the field of SHM.

	\section*{Acknowledgements}

	The authors wish to gratefully acknowledge support for this work through grants from the Engineering and Physical Sciences Research Council (EPSRC), UK, and Natural Environment Research Council, UK via grant number, EP/S023763/1. For the purpose of open access, the author has applied a Creative Commons Attribution (CC BY) licence to any Author Accepted Manuscript version arising.
	L.A.\ Bull was supported by Wave 1 of The UKRI Strategic Priorities Fund under the EPSRC Grant EP/W006022/1, particularly the \textit{Ecosystems of Digital Twins} theme within that grant and The Alan Turing Institute.

	\bibliography{14380_Bee_library.bib}
	\bibliographystyle{ieeetr}
	
\end{document}